\documentclass[conference]{IEEEtran}
\IEEEoverridecommandlockouts
\usepackage{cite}
\usepackage{amsmath,amssymb,amsfonts}
\usepackage{algorithmic}
\usepackage{graphicx}
\usepackage{textcomp}
\usepackage{xcolor}
\def\BibTeX{{\rm B\kern-.05em{\sc i\kern-.025em b}\kern-.08em
    T\kern-.1667em\lower.7ex\hbox{E}\kern-.125emX}}
\begin{document}

\title{GLoG-CSUnet: Enhancing Vision Transformers with Adaptable Radiomic Features for Medical Image Segmentation\\

}

\author{\IEEEauthorblockN{Niloufar Eghbali Zarch}
\IEEEauthorblockA{\textit{Computer Science Department} \\
\textit{Michigan State University}\\
East Lansing, USA \\
eghbaliz@msu.edu}
\and 
\IEEEauthorblockN{Hassan Bagher-Ebadian}
\IEEEauthorblockA{\textit{  Department of Radiation Oncology
} \\
\textit{Henry Ford Health}\\
Detroit, US \\
 hbagher1@hfhs.org
}
\and

\IEEEauthorblockN{Tuka Alhanai}
\IEEEauthorblockA{\textit{Division of Engineering} \\
\textit{New York University}\\
Abu Dhabi, UAE \\
tuka.alhanai@nyu.edu}
\and
\IEEEauthorblockN{Mohammad M. Ghassemi}
\IEEEauthorblockA{\textit{Computer Science Department} \\
\textit{Michigan State University}\\
East Lansing, USA \\
ghassem3@msu.edu}

}

\maketitle

\begin{abstract}
Vision Transformers (ViTs) have shown promise in medical image semantic segmentation (MISS) by capturing long-range correlations. However, ViTs often struggle to model local spatial information effectively, which is essential for accurately segmenting fine anatomical details, particularly when applied to small datasets without extensive pre-training.  We introduce Gabor and  Laplacian of Gaussian Convolutional Swin Network (GLoG-CSUnet), a novel architecture enhancing Transformer-based models by incorporating learnable radiomic features. This approach integrates dynamically adaptive Gabor and Laplacian of Gaussian (LoG) filters to capture texture, edge, and boundary information, enhancing the feature representation processed by the Transformer model.
Our method uniquely combines the long-range dependency modeling of Transformers with the texture analysis capabilities of Gabor and LoG features. Evaluated on the Synapse multi-organ and ACDC cardiac segmentation datasets, GLoG-CSUnet demonstrates significant improvements over state-of-the-art models, achieving a 1.14\% increase in Dice score for Synapse and 0.99\% for ACDC, with minimal computational overhead (only 15 and 30 additional parameters, respectively). GLoG-CSUnet's flexible design allows integration with various base models, offering a promising approach for incorporating radiomics-inspired feature extraction in Transformer architectures for medical image analysis. The code implementation is available on GitHub at: https://github.com/HAAIL/GLoG-CSUnet.
\end{abstract}


\section{Introduction}

\noindent Medical image semantic segmentation (MISS) is a critical task in healthcare, enabling precise identification and delineation of anatomical structures including organs and lesions. MISS is essential for accurate diagnosis, treatment planning, and disease monitoring \cite{daza2021towards, wang2022medical}. The accuracy of MISS is particularly important in applications such as radiotherapy and surgical planning, where precision directly impacts patient outcomes \cite{zaidi2010pet, gering2001integrated}.

The significance of MISS has driven continuous research into advanced segmentation methods. Initially, convolutional neural networks (CNNs), especially U-shaped architectures with skip-connections like U-Net, demonstrated high effectiveness in medical image segmentation \cite{ronneberger2015u,zhou2018unet++,huang2020unet}. However, these models struggle with the intrinsic locality of convolution operations, limiting their ability to model global context effectively \cite{chen2021transunet}.

Vision Transformers (ViTs) have recently emerged as powerful alternatives to CNNs for MISS \cite{chen2021transunet}. The ViT model is designed to capture long-range dependencies within an image by splitting it into patches and treating each patch as a sequence element, similar to how words are processed in natural language processing models \cite{dosovitskiy2020image}. Unlike CNNs, which focus on local pixel relationships, ViTs use self-attention mechanisms to model global interactions between image patches, making them particularly effective for tasks where understanding global context is crucial \cite{cao2022swin}. 
This ability to model long-range dependencies between image patches allows for more global context awareness, which is particularly useful for segmenting complex structures across medical images\cite{dosovitskiy2020image,chen2021transunet}.

\subsection{Related Works}
\noindent Several studies have explored hybrid approaches to medical image segmentation and report improved performance over traditional CNN-based methods; notable examples include: (i) TransUNet \cite{chen2021transunet}, which combines CNNs and Transformers, and (ii) Swin-Unet \cite{cao2022swin}, a fully Transformer-based model within a CNN-based framework. Inspired by the success of Swin-Unet, several extensions have been proposed to further improve its performance, particularly in terms of local feature extraction and computational efficiency. Wang et al. \cite{wang2022mixed},  introduced a novel Mixed Transformer Module (MTM), combining Local-Global Gaussian-Weighted Self-Attention (LGG-SA) and External Attention (EA) to capture both local and global dependencies. This model focuses on reducing computational complexity while maintaining high accuracy by emphasizing nearby patches more effectively.
Similarly, Liu et al. \cite{liu2023optimizing} extended Swin-Unet by incorporating Convolutional Swin Transformer (CST) blocks, which add convolutional layers within the Transformer architecture to enhance spatial and local feature modeling. This approach addresses one of the limitations of purely Transformer-based models, which often struggle with boundary refinement and local feature extraction. By merging convolutions with multi-head self-attention, CST achieves better segmentation performance, especially on small datasets like those commonly used in medical image segmentation.

\subsection{Challenges and Our Contribution}

\noindent Despite their potential, ViTs have limitations in capturing local spatial information, especially when applied to small datasets without extensive pre-training. 
This makes it difficult for them to accurately segment fine details such as the boundaries of small organs, which are critical in medical image analysis. Moreover, combining both CNNs and Vision Transformers (ViTs) in hybrid models can lead to significantly large and computationally expensive networks, further presenting a challenge in resource-constrained environments.

To address these challenges, we propose GLoG-CSUnet, a novel architecture that enhances ViT-based models by integrating learnable radiomic features, specifically Gabor filters and Laplacian of Gaussian (LoG) filters, into the Transformer framework. Radiomics, the high-throughput extraction of quantitative features from medical images, has demonstrated great potential in improving diagnostic and predictive accuracy across various clinical applications \cite{mayerhoefer2020introduction,depeursinge2020standardised}. Among radiomic features, Gabor filters are particularly effective at capturing multi-scale and multi-orientation texture information, which is crucial for characterizing complex tissue structures in medical images \cite{ardakani2022interpretation}. LoG filters, on the other hand, are valuable for improving edge detection and boundary precision, which are key to accurately segmenting anatomical structures \cite{depeursinge2020standardised}.

 Our approach leverages the strength of Gabor and LoG filters in extracting these discriminative texture and edge features, which complement the global context modeling capabilities of ViTs. By integrating these adaptive local radiomic features with the transformer architecture, GLoG-CSUnet aims to achieve a better balance between local detail preservation and global context understanding. This integration is particularly beneficial for smaller datasets where extensive pre-training is not feasible, as it enhances the model's ability to capture fine-grained anatomical details without relying solely on large-scale data for feature learning.

\begin{figure}[!ht]
\includegraphics[width=\linewidth]{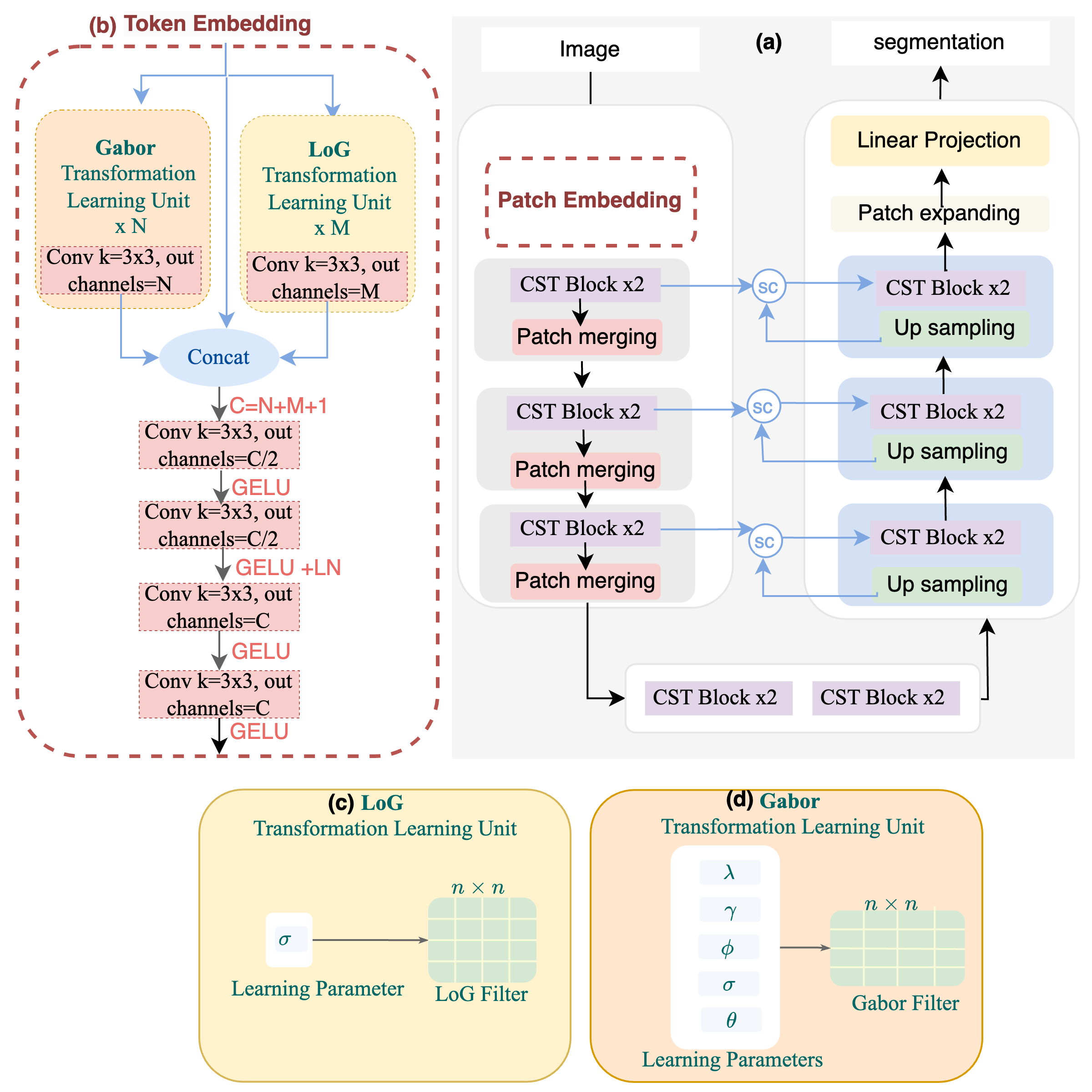}
\caption{An overview of the GLoG-CSUnet model architecture. (a) Segmentation Pipeline: the image is first processed through the Patch Embedding module, followed by multiple Convolutional Swin Transformer (CST) blocks. (b) Patch Embedding: transforms the input image into a format suitable for processing by the transformer blocks. It includes Gabor and Laplacian of Gaussian (LoG) transformation learning units. The outputs of these units are concatenated and further processed through multiple convolutional layers with Gaussian Error Linear Unit (GELU) activation and layer normalization (LN) to prepare the embedded Patches.(c) LoG Transformation Learning Unit:  This unit learns the LoG filter parameters (d) Gabor Transformation Learning Unit: This unit learns the Gabor filter parameters.}
\label{overview}
\end{figure}
\section{Method}

\noindent In this section, we describe the architecture and components of GLoG-CSUnet. An overview of the method is presented in \ref{overview}. In addition, we detail the key elements of our approach, including the adaptable Gabor and LoG filters, data preparation, evaluation metrics, and experimental setup.

\subsection{Model Overview}
\noindent GLoG-CSUnet extends the Convolutional Swin-Unet (CSUnet) architecture \cite{liu2023optimizing}, by incorporating adaptive Gabor filters and LoG filters into the segmentation pipeline. The architecture consists of the following main components:
\noindent\begin{itemize}
    \item {\textbf{Gabor transformation Layer:}} Utilizes learnable Gabor filters to extract multi-scale and multi-orientation texture and edge information. These filters dynamically adapt during training to optimize the capture of radiomic features.
    \item{\textbf{LoG  transformation Layer:}} Employs LoG filters to enhance boundary precision through edge detection and improved boundary delineation. The LoG filter parameters are optimized during the training process.

    \item{\textbf{Convolutional Patch Embedding Layer}}  divides the input images along with radiomics transformations into non-overlapping patches, through a series of convolutional layers, preserving spatial locality before passing the patches to the Transformer.
    \item {\textbf{Vision Transformer Backbone:} }Swin Transformer serves as the core of GLoG-CSUnet, operating on image patches through shifted windows. This Transformer block models long-range dependencies, capturing global context, while Gabor and LoG filters handle local feature extraction.
\end{itemize}

\subsection{Adaptable Filters}
\subsubsection{Gabor Filter}
Gabor filters are particularly effective at extracting texture information and directional features from images. In our GLoG-CSUnet model, we implement a learnable Gabor filter layer that adapts to the specific characteristics of medical images during training.
The 2D Gabor function \cite{granlund1978search} is defined by the following equation:

\begin{equation}
\text{Gabor}(x_{\theta}, y_{\theta}) = e ^{\left( -\frac{x_{\theta}^2 + \gamma^2 y_{\theta}^2}{2\sigma^2} \right)}  \cos\left( 2\pi \frac{x_{\theta}}{\lambda} + \psi \right)
\end{equation}
where  
$x_{\theta}=x\cos{\theta}+y\sin{\theta}$ and $y_{\theta}=-x\sin{\theta}+y\cos{\theta}$. In this equation $\lambda$ is the wavelength of sinusoidal component, $\theta $ is the orientation of the normal to the parallel stripes of the Gabor function, $\psi$ is the phase offset, $\sigma $ is the standard deviation of the Gaussian envelope and $\lambda$  is the spatial aspect ratio, determining the ellipticity of the Gabor function. In our implementation, these parameters are learned during the training process.

\subsubsection{LoG Filter}
The Laplacian of Gaussian (LoG) filter is used to detect edges and capture fine anatomical details by identifying areas of rapid intensity change. The LoG filter is represented as:
\begin{equation}
LoG(x, y) = -\frac{1}{\pi \sigma^4} \left(1 - \frac{x^2 + y^2}{2\sigma^2}\right) e ^{\left(-\frac{x^2 + y^2}{2\sigma^2}\right)}
\end{equation}
where $\sigma$is the scale parameter that controls the sensitivity of the filter to edge detection at different scales. The LoG filter operates by first smoothing the image with a Gaussian filter and then applying the Laplacian operator to highlight regions of rapid intensity change. In GLoG-CSUnet, the LoG filters are implemented as learnable filters with a trainable $\sigma$ parameter. 
\subsection{Data}
In this study, we utilized two publicly available datasets to evaluate the performance of our segmentation model.
\subsubsection{Synapse Dataset \cite{landman2015miccai}}
The Synapse multi-organ segmentation dataset consists of contrast-enhanced abdominal CT scans from 30 subjects, covering 8 organs: aorta, gallbladder, spleen, left kidney, right kidney, liver, pancreas, and stomach. Following the standard split from previous works \cite{wang2022mixed,liu2023optimizing,chen2021transunet}, we used 18 cases (2212 axial slices) for training and 12 cases for testing. Each image in the dataset has corresponding ground truth annotations for the organs. The images were resampled to a resolution of $224\times224$ ACDC Dataset to match the input requirements of our network. To enhance the model's robustness, we applied data augmentation techniques, including random flipping and rotation.
\subsubsection{ACDC Dataset \cite{bernard2018deep} }
The Automatic Cardiac Diagnosis Challenge (ACDC) dataset contains MRI images from 100 patients, focusing on the segmentation of three cardiac structures: the right ventricle (RV), left ventricle (LV), and myocardium (MYO). The dataset is split into 70 exams (1930 axial slices) for training, 10 exams for validation, and 20 exams for testing, following the protocol established in previous studies \cite{wang2022mixed,liu2023optimizing}. The images were resized to $224\times224$ for consistency across datasets, and augmentation techniques such as flipping and rotation were applied to the training data, similar to approaches used in previous studies.

\subsection{Evaluation}
\noindent Following the prior works, we employed two key metrics to assess the performance of our model and the baselines: the Dice Similarity Coefficient (DSC) \cite{taha2015metrics} and the 95th percentile Hausdorff Distance (HD95) \cite{huttenlocher1993comparing}. The DSC, a widely adopted metric in medical image segmentation, quantifies the overlap between predicted and ground-truth segmentations; it ranges from 0 to 1, with 1 indicating perfect overlap. The DSC is particularly valuable for evaluating segmentation accuracy in scenarios involving imbalanced datasets or small structures \cite{taha2015metrics}. The HD95 measures spatial accuracy by calculating the maximum distance between boundary points of the predicted and ground truth segmentation, using the 95th percentile to reduce outlier effects. 

\subsection{Experimental Setup}
 \noindent Our proposed model was trained from scratch for 300 epochs, starting from a randomly initialized set of weights. We used a batch size of 24 and optimized the model using the AdamW optimizer \cite{loshchilov2017decoupled} with a weight decay of 2e-4 across both datasets. We applied dataset-specific learning rates: 1e-3 for the Synapse dataset and 1e-2 for the ACDC dataset.  training dynamics. 
For the ACDC dataset, we trained 5 Gabor and 5 LoG filters, while for the Synapse dataset, we used 2 Gabor and 5 LoG filters.  
For baseline models, pre-trained weights were used if provided.

\section{Results}

\noindent We evaluated GLoG-CSUnet against several state-of-the-art models on both the Synapse and ACDC datasets. Table \ref{table:synapse-results} presents the performance comparison on the Synapse multi-organ segmentation dataset. Our GLoG-CSUnet model achieved the highest overall Dice Similarity Coefficient (DSC) of 83.36\%, surpassing the previous best performance of 82.21\% by CS-Unet. GLoG-CSUnet demonstrated superior performance in six out of eight organ categories, with notable improvements in challenging organs such as the gallbladder (73.20\% DSC) and stomach (82.10\% DSC). While our model's D of 23.02 was not the lowest, it still represented a significant improvement over all baseline methods except Swin-Unet, indicating good boundary delineation capabilities.

\begin{table*}[]
\centering
\caption{Experimental results of the Synapse Dataset. The Dice Similarity Coefficient (DSC) and Hausdorff Distance (HD) for each single organ class are presented.}
\resizebox{\textwidth}{!}{

\begin{tabular}{l|c|c|c|c|c|c|c|c|c|c}
\hline
\textbf{Methods} & \textbf{Overall DSC} & \textbf{Overall HD} & \textbf{Aorta} & \textbf{Gallb} & \textbf{Kid\_L} & \textbf{Kid\_R} & \textbf{Liver} & \textbf{Pancr} & \textbf{Spleen} & \textbf{Stom} \\ \hline
R50 UNet \cite{chen2021transunet}     & 74.68        & 36.87       & 84.18          & 62.84          & 79.19          & 71.29          & 93.35          & 48.23          & 84.41           & 73.92         \\ 
R50 AttnUNet \cite{chen2021transunet} & 75.57        & 36.97       & 55.92          & 63.91          & 79.20          & 72.71          & 93.56          & 49.37          & 87.19           & 74.95         \\
 \hline
R50 ViT \cite{chen2021transunet}      & 71.29        & 32.87       & 73.73          & 55.13          & 75.80          & 72.20          & 91.51          & 45.99          & 81.99           & 73.95         \\ 
TransUNet \cite{chen2021transunet}    & 77.48        & 31.69       & 87.23          & 63.13          & 81.87          & 77.02          & 94.08          & 55.86          & 85.08           & 75.62         \\ 
Swin-UNet \cite{cao2022swin}  & 79.13        & \textbf{21.55  }     & 85.47          & 66.53          & 83.28          & 79.61          & 94.29          & 56.58          & 90.66           & 76.60         \\ 
MT-UNet \cite{wang2022mixed}    & 78.59        & 26.59       & 87.92          & 64.99          & 81.47          & 77.29          & 93.06          & 59.46          & 87.75           & 76.81         \\ 
CS-Unet \cite{liu2023optimizing}             & 82.21        & 27.02       & 88.40& 72.59          & \textbf{85.28}          & 79.52          & 94.35          & 70.12          & 91.06           & 75.72         \\ \hline
Our Method & \textbf{83.36}&23.02&\textbf{88.43}&\textbf{73.20}&84.50&\textbf{80.98}&\textbf{95.14}&\textbf{71.02}&\textbf{91.51}&\textbf{82.10} \\ \hline
\end{tabular}%

}
\label{table:synapse-results}
\end{table*}

Table \ref{table:acdc_results} shows the results for the ACDC cardiac segmentation dataset. GLoG-CSUnet achieved the highest overall DSC of 92.28\%, outperforming the previous best model by 0.91 percentage points. Our model demonstrated superior performance across all three cardiac structures: Right Ventricle (91.04\% DSC), Myocardium (90.09\% DSC), and Left Ventricle (95.71\% DSC). These results, consistent across both datasets, highlight GLoG-CSUnet's effectiveness in capturing both local and global features for accurate medical image segmentation, particularly in challenging cases such as small organs and complex structures.

\begin{table}[h]
\caption{Experimental results of the ACDC Dataset.}
\centering
\resizebox{\columnwidth}{!}{
\begin{tabular}{lcccc}
\textbf{Method} & \textbf{DSC(\%)} & \textbf{RV} & \textbf{Myo} & \textbf{LV} \\
\hline
R50 UNet \cite{chen2021transunet} & 87.60 & 84.62 & 84.52 & 93.68 \\
R50 AttnUNet \cite{chen2021transunet} & 86.90 & 83.27 & 84.33 & 93.53 \\ 
\hline 

R50 ViT \cite{chen2021transunet} & 86.19 & 82.51 & 83.01 & 93.05 \\
TransUNet \cite{chen2021transunet} & 89.71 & 86.67 & 87.27 & 95.18 \\
Swin-Unet \cite{cao2022swin} & 88.07 & 85.77 & 84.42 & 94.03 \\
MT-Unet  \cite{wang2022mixed} & 90.43 & 86.64 & 89.04 & 95.62 \\
CS-Unet  \cite{liu2023optimizing} & 91.37 & 89.20 & 89.47 & 95.22 \\ \hline
Our Method &\textbf{92.28}&\textbf{91.04}&\textbf{90.09}&\textbf{95.71}\\
\hline
\end{tabular}
\label{table:acdc_results}
}
\end{table}

\begin{figure}[!h]
\centering
\includegraphics[width=\columnwidth]{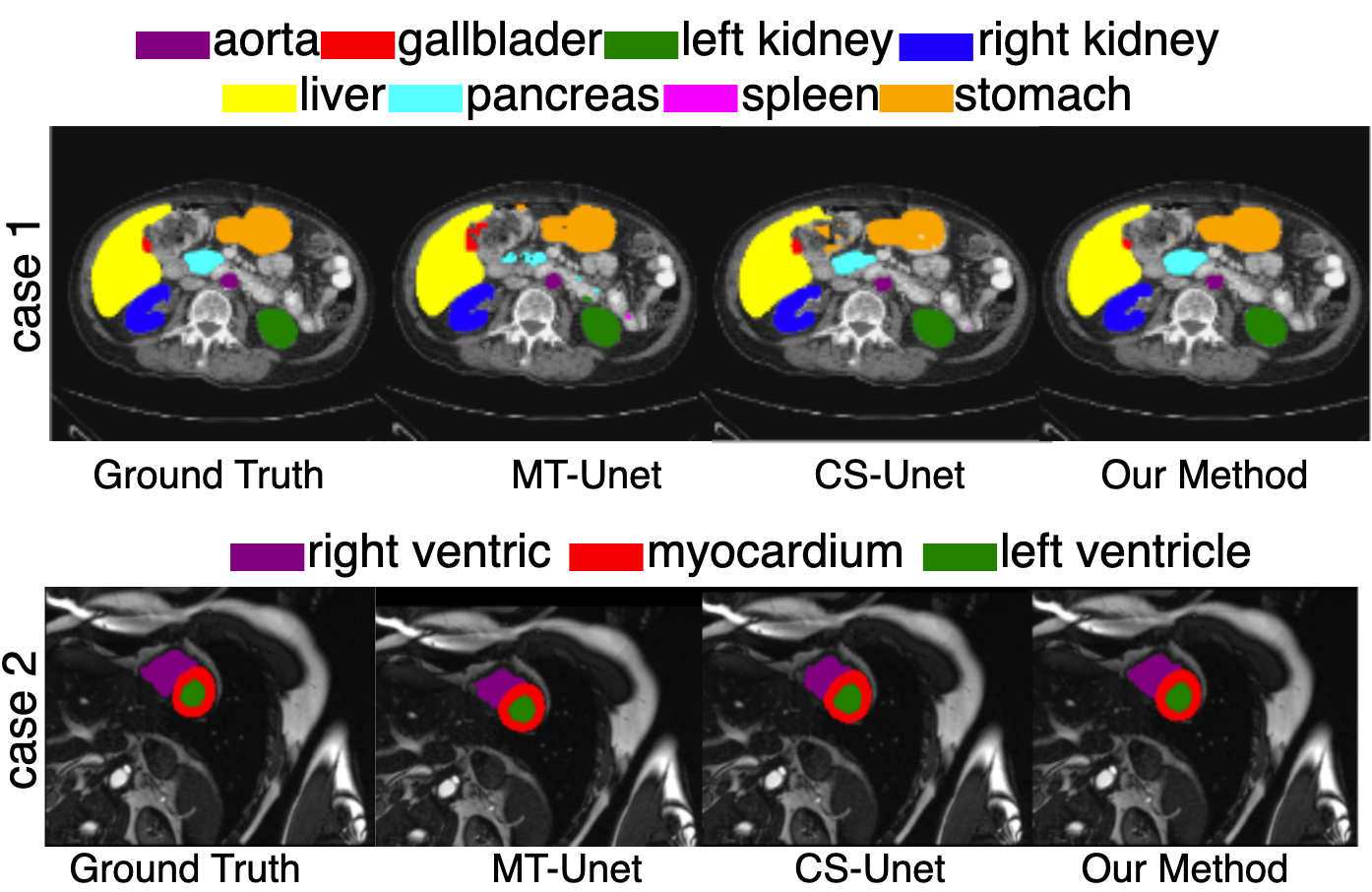}
\caption{ Visualization of segmentation results for two test cases using our method alongside the MT-Unet and CS-Unet baseline methods. case 1 and case 2 were sampled from Synapse and ACDC datasets respectively. }
\label{visualizan}
\end{figure}

Figure \ref{visualizan} displays the segmentation results for two test cases, comparing our method against the top two baseline methods, MT-Unet and CS-Unet. In Case 1, our method achieves precise segmentation across all organs, notably outperforming other methods in accurately delineating the stomach and pancreas. In Case 2, although all methods effectively segment the left ventricle and myocardium, our model excels in segmenting the right ventricle.

\subsection{Ablation Study}

To validate the effectiveness of each component in our architecture, we conducted an ablation study comparing three variants: Gabor-only CSUnet, LoG-only CSUnet, and the complete GLoG-CSUnet using the ACDC dataset. Results showed that while both filter types independently improved performance, their combination yielded the best results. The Gabor-only variant achieved a DSC of 92.01\%, while the LoG-only version reached 91.83\%. The complete GLoG-CSUnet demonstrated superior performance with a DSC of 92.28\%.


\section{Discussion}

\noindent The results demonstrate that GLoG-CSUnet consistently outperforms state-of-the-art methods across both the Synapse multi-organ and ACDC cardiac segmentation datasets. This is achieved with only a minimal increase in model parameters—adding just 15 parameters for the Synapse and 30 for the ACDC dataset, to the full model's total of 24 million parameters. This performance improvement, achieved with such a small computational overhead, can be attributed to the integration of learnable radiomic features, specifically Gabor and LoG filters, with the Transformer architecture. The superior performance in segmenting challenging structures, particularly the stomach, underscores the effectiveness of our approach in capturing fine-grained local details. Our GLoG-CSUnet model achieved a Dice Similarity Coefficient (DSC) for stomach segmentation of 82.10\%, which represents a substantial improvement of 5.29\% over the next best performance by MT-Unet. This significant increase, along with consistent enhancements across other structures, indicates the sucsses of our method in blending detailed local feature extraction with effective global context modeling. The adaptive nature of our Gabor and LoG filters allows the model to optimize feature extraction for each specific segmentation task, addressing a key limitation of standard Transformer-based models in capturing local spatial information.

 Future work could explore the interpretability of the learned filters, potentially providing insights into the specific image features most relevant for different anatomical structures. 

\section{Conclusion}
\noindent In conclusion, GLoG-CSUnet presents a advancement in medical image segmentation, effectively combining the strengths of radiomics-inspired feature extraction with the global context modeling capabilities of Transformers. This approach not only improves segmentation accuracy with minimal computational overhead but also opens new avenues for integrating domain-specific knowledge into deep learning models for medical image analysis.

\begingroup
\bibliographystyle{icassp24}  
\bibliography{icassp24}    
\endgroup

\end{document}